\def\carearxiv{}
\def\carecodelink{}

\documentclass[10pt,twocolumn,letterpaper]{article}

\ifdefined\carearxiv
  \usepackage[pagenumbers]{cvpr} %
\else
  \usepackage[review]{cvpr}      %
\fi

\usepackage[utf8]{inputenc}
\usepackage{multirow}
\usepackage{array,longtable,tabularx}
\usepackage{placeins}

\definecolor{rowgray}{gray}{0.93}
\definecolor{ourscolor}{RGB}{30,78,140}
\definecolor{lewmcolor}{RGB}{176,125,44}
\definecolor{ahccolor}{RGB}{120,120,120}

\makeatletter
\newcommand{\pmstd}[1]{%
  {\color{black!60}\fontsize{\dimexpr\f@size pt*4/5\relax}{\f@baselineskip}%
   \selectfont$\pm$#1}}
\makeatother
\newcommand{\res}[2]{#1\pmstd{#2}}
\newcommand{\resb}[2]{\textbf{#1}\pmstd{#2}}
\newcommand{\resu}[2]{\underline{#1}\pmstd{#2}}
\newcommand{\oursrow}{\rowcolor{rowgray}}
\newcommand{\nores}{--}
\newcommand{\phz}{\phantom{0}}
\newcommand{\phm}{\phantom{$-$}}
\newcolumntype{C}{>{\centering\arraybackslash}X}
\newcolumntype{L}{>{\raggedright\arraybackslash}X}

\newcommand{\pred}{\mathrm{pred}}

\newcommand{\name}{CARE-WM}

\newcommand{\horizonfigurecaptionspace}{-1.5em}
\newenvironment{conclusionblock}
  {\ifdefined\carearxiv\else\samepage\fi}
  {\par}

\usepackage{algorithm}
\usepackage[noend]{algorithmic}
\usepackage{float}
\usepackage{colortbl}

\definecolor{cvprblue}{rgb}{0.21,0.49,0.74}
\usepackage[pagebackref,breaklinks,colorlinks,allcolors=cvprblue]{hyperref}
\usepackage{orcidlink} %

\def\confName{CVPR}
\def\confYear{2027}

\ifdefined\carecodelink
  \newcommand{\codeavailability}{%
    The code is available at
    {\urlstyle{same}\url{https://github.com/KaidRommel/CARE-WM}}.}
\else
  \newcommand{\codeavailability}{}
\fi

\title{%
  CARE-WM: Same-Horizon Self-Conditioning with Constraint-Adaptive Regularization for Latent World Models%
}
\author{%
Yuchen Lin\,\orcidlink{0009-0003-0999-3504}\\
University of Southern California\\
{\tt\small ylin7261@usc.edu}
\and
Kai Zhao\,\orcidlink{0000-0002-2496-0829}%
\thanks{Corresponding author.}\\
Shanghai University\\
{\tt\small kz@kaizhao.net}
}

\begin{document}
\maketitle

\begin{abstract}
Planning with latent world models requires both accurate multi-step predictions
and latent distances that reflect physical proximity to the goal.
However, models are typically trained on ground-truth observations but must rely
on their own imaginations during planning, while closer latent distances do not
necessarily correspond to better task outcomes.
Existing methods resort to auxiliary training objectives or additional
task-cost readouts, which respectively incur extra training overhead or leave
the latent geometry itself misaligned with task cost.
We introduce \name, a drop-in training method that meets both requirements without
additional supervision, a separate cost model, or any change to LeWM's architecture
and Euclidean-cost planner.
\name{} trains the predictor on its own predictions and adaptively calibrates
the strength of latent regularization.
With a single hyperparameter configuration, \name{} improves mean planning
success over LeWM across all evaluated continuous-control tasks and matches
or exceeds the best results of other LeWM-family methods on most tasks.
Further analyses show lower open-loop latent prediction error and
closer agreement between latent-cost rankings and physical task costs.
\codeavailability

\end{abstract}

\section{Introduction}\label{sec:intro}

World models support planning by predicting the consequences of candidate
actions~\cite{assran2025vjepa2selfsupervisedvideo,ha2018world,hafner2025mastering}.
LeWorldModel (LeWM)~\cite{maes_lelidec2026lewm} jointly learns visual
representations and action-conditioned latent dynamics from pixels,
without reconstructing them~\cite{lecun2022path}.
At planning time, the predictor recursively rolls out future latents, and
the cross-entropy method
(CEM)~\cite{rubinstein1999cross,zhou2025dino-wm} ranks candidate action
sequences by the Euclidean distance between their predicted terminal
latents and the encoding of a goal image. The model must both predict where
an action sequence will lead and judge how close that outcome comes to the
goal.
\begin{figure}[t]
  \centering
  \includegraphics[width=0.95\linewidth]{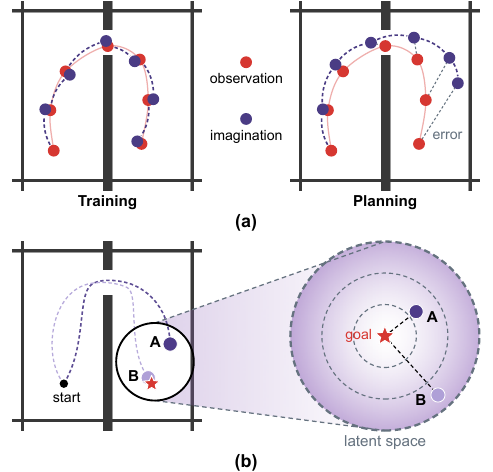}
  \caption[Two ways latent-space planning can fail.]{%
    \input{shared/teaser-caption.tex}}
  \label{fig:care-teaser}
\end{figure}

Planning can fail in two distinct ways as illustrated in~\cref{fig:care-teaser}.
First, a predictor that is accurate on observed histories may become unreliable
when it repeatedly uses its own predictions as inputs~\cite{bengio2015scheduled,venkatraman2015improving}. During training,
teacher forcing provides encoded observations, whereas planning feeds
predictions back into the model. Errors can propagate and alter candidate rankings.
We refer to accurate multi-step prediction under such model-generated inputs
as \emph{rollout accuracy}.

Second, even a correctly predicted endpoint can receive a misleading planning cost.
The Euclidean distance between its latent representation and the goal may not
reflect the actual task outcome~\cite{li2026predictive,wang2026decisionmetric}.
For example, in a two-room environment, an endpoint separated from the goal by
a wall may appear closer than one along a feasible route. We refer to the agreement
between latent distances at executed endpoints and
their physical task costs as \emph{planning-cost alignment}.

Existing methods address rollout error by supervising latent predictions
at multiple depths of recursive rollouts~\cite{hafner2019planet,hansen2024tdmpc2,li2026predictive}.
Other approaches replace recursive rollout with direct prediction
of future states conditioned on the initial observation and action
sequence~\cite{gao2026fast-lewm,li2026beyond}.
On the cost side, reachability and decision-related objectives give the
representation information about which futures are attainable or
preferable~\cite{li2026predictive,wang2026decisionmetric,liu2026djepa, yan2026forward}.
These objectives can use the original trajectories while adding
supervision beyond next-latent prediction.
This motivates our central question.
\emph{Can we improve both aspects of planning reliability through
training alone, without additional task supervision or changes to the
prediction and planning interfaces?}

We introduce \name{}, a training method with two components for improving
rollout accuracy and planning-cost alignment.
We hypothesize that exposure bias from the mismatch between observed
training and generated rollout contexts~\cite{bengio2015scheduled,ross2011dagger}
is a key source of recursive error.
The first component, Same-Horizon Self-Conditioning (SHSC), therefore trains
the predictor on its own outputs as inputs and backpropagates through them
within the original supervision window.
To improve planning-cost alignment without task-specific objectives, we
seek to relax unnecessary regularization constraints~\cite{tandon2026localejepa}.
LeWM uses Sketched Isotropic Gaussian Regularization
(SIGReg)~\cite{balestriero2025lejepa} to prevent collapse, yet we find that
finite-sample bias sustains its regularization pressure even at the
Gaussian target.
Manually tuning the regularization weight to temper this pressure
requires additional hyperparameter search.
\name{}'s second component, Constraint-Adaptive SIGReg (CA-SIGReg),
therefore adjusts this weight automatically using a finite-sample-calibrated
discrepancy from the Gaussian target.

\name{} improves mean planning success over LeWM across all evaluated
continuous-control tasks and matches or exceeds the best results of other
LeWM-family methods on most tasks.
It achieves lower open-loop latent MSE at every
evaluated horizon on all four tasks in the rollout evaluation.
Its predicted costs better
preserve realized latent-cost rankings, and realized costs rank candidates
more consistently with physical task costs.
Our contributions are threefold.
\begin{itemize}
    \item \textbf{Same-Horizon Self-Conditioning (SHSC).}
    We introduce a training scheme that improves rollout accuracy by exposing the
    predictor to its own predictions without extending the supervision horizon.
    \item \textbf{Constraint-Adaptive SIGReg (CA-SIGReg).}
    We propose an adaptive regularization strategy that balances collapse prevention
    with the flexibility to learn planning-relevant latent geometry.
    \item \textbf{Empirical evaluation of planning reliability.}
    We demonstrate improved planning success across five continuous-control tasks
    and investigate both rollout accuracy and planning-cost alignment through
    targeted analyses and component ablations.
\end{itemize}

\section{Related Work}\label{sec:related}

\paragraph{Visual latent world models for planning.}
Generative world models learn latent dynamics with observation
reconstruction~\cite{ha2018world,hafner2019planet,hafner2025mastering},
whereas JEPAs predict target representations
directly~\cite{lecun2022path,assran2023self,bardes2024revisiting}.
For goal-conditioned planning, DINO-WM~\cite{zhou2025dino-wm} learns
dynamics over frozen pretrained visual features, whereas
PLDM~\cite{sobal2025pldm} and LeWM~\cite{maes_lelidec2026lewm} learn the
encoder and predictor jointly from pixels. We build on
LeWM to study how training affects planning with its original encoder,
predictor, and Euclidean-cost CEM planner.

\paragraph{Learning under model-generated contexts.}
Exposure bias~\cite{bengio2015scheduled,ranzato2015sequence}, the mismatch
between observed training inputs and generated test-time inputs, has
motivated training on generated inputs.
Scheduled sampling mixes observed and generated inputs during sequence
training~\cite{bengio2015scheduled}, while DAgger queries an expert at
states visited by the learner~\cite{ross2011dagger}. For learned dynamics,
Data as Demonstrator reuses existing trajectories to supply targets for
model-generated inputs~\cite{venkatraman2015improving}. Hallucinated replay
pairs simulated states with observed next-state targets along the same
action sequence~\cite{talvitie2017self}.

Latent overshooting and TD-MPC2 supervise predictions at multiple rollout
depths~\cite{hafner2019planet,hansen2024tdmpc2}. RC-aux combines
multi-horizon open-loop prediction with reachability
supervision~\cite{li2026predictive}. Fast-LeWM predicts future latents in
parallel from the current encoding and action prefixes, avoiding repeated
application of a local transition~\cite{gao2026fast-lewm}. SHSC introduces
generated-input supervision in a jointly learned visual latent space. A
second causal call uses the same predictor on shifted teacher predictions
and their aligned actions. The student loss backpropagates through those
inputs to the teacher call. All targets lie within the original trajectory
window, and planning uses the same autoregressive prediction interface.

\paragraph{Planning-aware representations and regularization.}
Planning-aware objectives give the encoder information about which
outcomes are attainable or preferable. RC-aux uses trajectory offsets and
temporal hard negatives to supervise budget-conditioned
reachability~\cite{li2026predictive}. DA-LeWM adds inverse-dynamics and
demonstration-conditioned goal-action heads~\cite{wang2026decisionmetric}.
D-JEPA learns decision relations among candidate futures from their
executed outcomes~\cite{liu2026djepa}. Existing
planning-aware approaches explicitly introduce task- or decision-relevant
supervision, whereas \name{} investigates what can be achieved through
predictive supervision and representation regularization alone.

Joint prediction admits a collapsed solution in which every observation
has the same encoding. VICReg constrains feature variance and
covariance~\cite{bardes2022vicreg}. LeJEPA's SIGReg compares projected
encodings with a Gaussian reference~\cite{balestriero2025lejepa}, while
Sub-JEPA imposes Gaussian constraints in low-dimensional
subspaces~\cite{zhao2026subjepa}. Under assumptions on Gaussian latents
and stationary additive-noise transitions, Klindt et
al.~\cite{klindt2026lejepawm} show that LeJEPA can recover the underlying
state up to a linear transformation at a population optimum. CA-SIGReg
retains LeWM's SIGReg statistic and adjusts its weight using a discrepancy
calibrated for finite batches. This feedback measures how far projected
encodings depart from the Gaussian reference.

\section{Planning Reliability in Latent World Models}\label{sec:background}

\subsection{End-to-End Latent Prediction and Planning}
\label{sec:background_prediction}

LeWM~\cite{maes_lelidec2026lewm} trains an encoder $f_\psi$ and a predictor
$\pred_\phi$ jointly from pixels. The encoder maps an observation $o_t$ to a
latent $z_t=f_\psi(o_t)\in\mathbb R^D$. The predictor maps a latent context
$c_t$ and its aligned actions $\mathbf a_t$ to the next latent,
\begin{equation}
    \hat z_{t+1}=\pred_\phi(c_t,\mathbf a_t).
    \label{eq:predictor}
\end{equation}
For a full observed window, $c_t=z_{t-K+1:t}$ and $\mathbf a_t=a_{t-K+1:t}$.
Under teacher forcing, the context comes from encoded observations and
the target is the next observation's encoding. A planning rollout starts
from the observed history and appends each prediction to the sliding
context as the action window advances.

Planning instead feeds predictions back. The context window is initialized
with $K$ encoded observations and thereafter slides over the model's own
predictions, while the actions follow a candidate sequence
$a^{(i)}_{t:t+H-1}$. With the goal latent $z_g=f_\psi(o_g)$, the predicted
terminal latent $\hat z^{(i)}_{t+H}$ gives the cost
\begin{equation}
    \hat C_i=\frac{1}{D}\big\|\hat z^{(i)}_{t+H}-z_g\big\|_2^2.
    \label{eq:cem_cost}
\end{equation}
CEM~\cite{rubinstein1999cross,zhou2025dino-wm} samples candidates, refits its
sampling distribution to those with the lowest $\hat C_i$, and executes the
resulting plan before replanning from the new observation.

\subsection{Rollout Accuracy \& Planning-Cost Alignment}
\label{sec:background_planning}

To examine this ordering, each candidate can also be executed from the same
simulator state. Let $z^{(i)}_{t+H}=f_\psi(o^{(i)}_{t+H})$ encode the
resulting observation, and let $s^{(i)}_{t+H}$ and $s_g$ denote the reached
and goal simulator states. The realized latent goal cost and the physical
task cost are
\begin{equation}
    C_i=\frac{1}{D}\big\|z^{(i)}_{t+H}-z_g\big\|_2^2,
    \qquad
    J_i=J\big(s^{(i)}_{t+H},s_g\big),
    \label{eq:diagnostic_costs}
\end{equation}
where $J$ is a task-specific cost and lower values indicate better
outcomes. The planner uses $\hat C_i$ alone. We use $C_i$ and $J_i$ to
evaluate its ranking.

For planning, rollout accuracy determines whether $\hat C_i$ preserves
the ordering of $C_i$. Let
$\epsilon_i=\|\hat z^{(i)}_{t+H}-z^{(i)}_{t+H}\|_2^2/D$ be the terminal
prediction error. The triangle inequality gives
$|\sqrt{\hat C_i}-\sqrt{C_i}|\leq\sqrt{\epsilon_i}$, and multiplying by
$\sqrt{\hat C_i}+\sqrt{C_i}\leq2\sqrt{C_i}+\sqrt{\epsilon_i}$ yields
\begin{equation}
    \big|\hat C_i-C_i\big|\leq\epsilon_i+2\sqrt{\epsilon_i C_i}.
    \label{eq:cost_gap}
\end{equation}
For a fixed positive terminal MSE, the bound grows with the distance of the
realized endpoint from the goal.
If the realized costs of two candidates differ by more than the sum of
their error bounds, their predicted costs preserve their order.

For a fixed start and goal, let $\rho$ denote Spearman correlation across
candidate sequences. \emph{Rollout accuracy} concerns accuracy under
generated contexts. Its consequence for planning is measured by
$\rho(\hat C,C)$, the agreement between predicted and realized latent
costs. Even with exact rollouts, CEM inherits the ordering assigned by
the encoder to the executed endpoints. \emph{Planning-cost alignment}
concerns whether the encoder assigns lower realized costs to better
physical outcomes, as measured by $\rho(C,J)$. We compare the ordering
available to CEM directly with the task ordering through $\rho(\hat C,J)$.
We compute these correlations within each start--goal pair and then
average them equally across starts, since the planner ranks candidate
action sequences for a fixed start and goal.
These two requirements motivate training interventions that improve
prediction under generated contexts while controlling the representation
constraints that shape latent-space learning.

\section{\name{}}\label{sec:method}

\subsection{Overview}\label{sec:method_overview}

\begin{figure*}[t]
\centering
\includegraphics{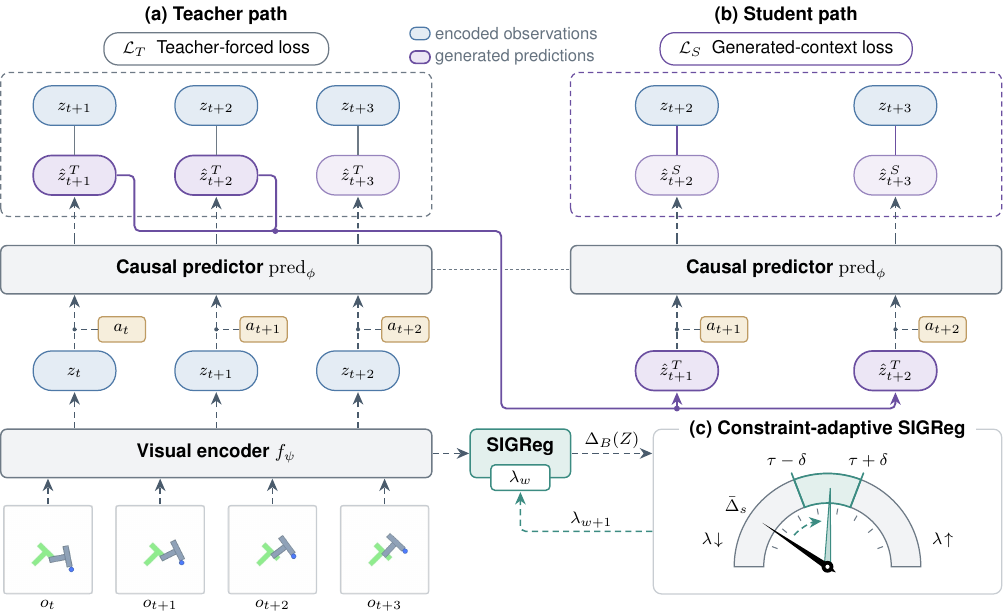}
\caption{\name{} training overview with a PushT observation window ($K=3$).
(a,b) SHSC applies the same predictor first to encoded observations and then
to its own predictions, preserving gradients through the reused inputs.
Both paths use supervision from the same window.
(c) CA-SIGReg adapts the strength of SIGReg using calibrated feedback from
the encodings. The black and teal needles show the current feedback signal
and the desired shift into the shaded tolerance band. Below the band,
$\lambda$ decreases; above it, $\lambda$ increases.}
\label{fig:care-overview}
\end{figure*}

As illustrated in \cref{fig:care-overview}, \name{} jointly trains
the encoder and predictor by combining SHSC's prediction losses with
the adaptively weighted SIGReg penalty from CA-SIGReg.
We first define the prediction losses,
then describe how the regularization weight is controlled, and finally
give the unified training objective.

\subsection{Same-Horizon Self-Conditioning}\label{sec:method_shsc}

SHSC adds a second, differentiable prediction pass conditioned on the first pass's outputs,
using only targets from the original training window.

\paragraph{Teacher and student prediction.}
For $K\geq2$, consider an encoded chunk $(z_{t:t+K},a_{t:t+K-1})$. A causal
teacher call predicts each next latent from observed prefixes, as in LeWM.
A student call of the same predictor takes the first $K-1$ teacher outputs
and their aligned actions as its context:
\begin{align}
    \hat z^{\,T}_{t+j+1}
    &=\pred_\phi\big(z_{t:t+j},a_{t:t+j}\big),
    \label{eq:teacher_path}\\
    \hat z^{\,S}_{t+j+1}
    &=\pred_\phi\big(\hat z^{\,T}_{t+1:t+j},a_{t+1:t+j}\big),
    \label{eq:student_path}
\end{align}
with $j=0,\ldots,K-1$ for the teacher and $j=1,\ldots,K-1$ for the student.
The final teacher output is supervised but not fed back. Both calls
therefore end at $z_{t+K}$, and the original $K+1$ observations supply
every target. Writing $\hat Z^T=\hat z^T_{t+1:t+K}$ and
$\hat Z^S=\hat z^S_{t+2:t+K}$, the losses are
\begin{equation}
\begin{aligned}
    \mathcal L_T
    &=\operatorname{MSE}\big(\hat Z^T,z_{t+1:t+K}\big),\\
    \mathcal L_S
    &=\operatorname{MSE}\big(\hat Z^S,z_{t+2:t+K}\big).
\end{aligned}
\label{eq:shsc_loss}
\end{equation}
Each MSE averages over its batch, prediction positions, and latent
coordinates, so $\mathcal L_T$ is LeWM's teacher-forced loss.

\paragraph{Differentiable generated feedback.}
Teacher and student are two calls to the same trainable predictor.
With detached teacher outputs, the student loss still trains the shared predictor to
predict accurately from generated latents. SHSC additionally backpropagates gradients of
the student loss through these latents into the teacher call, encouraging the predictor
to produce latents that support subsequent prediction.

\subsection{Constraint-Adaptive SIGReg}\label{sec:ca_reg}

CA-SIGReg adapts the weight of the existing SIGReg penalty using
a finite-sample-calibrated discrepancy from its Gaussian target.

\paragraph{Finite-sample calibration.}
Let $Z\in\mathbb R^{(K+1)\times B\times D}$ collect the encodings of a
minibatch of $B$ windows, and let $R(Z)$ be LeWM's raw SIGReg
statistic~\cite{balestriero2025lejepa,maes_lelidec2026lewm}.
Even for encodings drawn from the standard Gaussian target, finite-sample
fluctuations give $R(Z)$ a positive expectation.
We denote this expectation by $R_{\mathrm N}$, assuming independent
encodings across the batch at each time position.
Setting every encoding to the zero vector gives the zero-collapse
reference $R_{\mathrm C}(B)=R(0)$.
For our fixed training batch size, $R_{\mathrm C}(B)>R_{\mathrm N}>0$, and
the calibrated discrepancy is
\begin{equation}
    \Delta_B(Z)=\frac{R(Z)-R_{\mathrm N}}{R_{\mathrm C}(B)-R_{\mathrm N}}.
    \label{eq:ca_health}
\end{equation}
It has expectation zero at the Gaussian reference and equals one at zero
collapse. For an individual minibatch, $\Delta_B(Z)$ can fall outside $[0,1]$.

\paragraph{Adaptive coefficient update.}
The loss keeps the raw penalty $\lambda_w R(Z)$, where $w$ indexes
coefficient update windows. Let $\bar\Delta_s$ denote the exponential
moving average (EMA) of the calibrated discrepancy at optimizer step $s$,
used only as detached feedback. After warmup, the coefficient is updated
periodically. For discrepancy tolerance $\tau$, deadband half-width
$\delta\geq0$, and step size $\eta_w>0$, the update at the final step $s$
of window $w$ is
\begin{equation}
    \lambda_{w+1}
       =\operatorname{clip}\!\left(
         \lambda_w+\eta_w\,\operatorname{deadband}_{\delta}
         (\bar\Delta_s-\tau)\right).
\label{eq:ca_controller}
\end{equation}
The deadband term is zero when the feedback lies within
$[\tau-\delta,\tau+\delta]$. Outside this band, it returns the signed
distance to the nearest boundary, giving a positive update above the band
and a negative update below it. The clip operator restricts the coefficient
to preset bounds, and the coefficient stays fixed between updates.

\subsection{Unified Training Objective}\label{sec:method_total}

The student weight $\alpha_e$ at epoch $e$ increases linearly from zero
to one during warmup and stays at one thereafter. The full objective is
\begin{equation}
    \mathcal L_{\mathrm{CARE}}
    =\frac{\mathcal L_T+\alpha_e\mathcal L_S}{1+\alpha_e}
       +\lambda_w R(Z).
    \label{eq:total_loss}
\end{equation}

Due to space constraints, the EMA update rule, formal definitions of the
update operators, warmup schedules, training settings, and complete training
algorithm are provided in the supplementary material.

\section{Experiments}\label{sec:exp}

We first evaluate planning success, then examine predictive fidelity and
planning-cost fidelity. Component comparisons and mechanism diagnostics
assess the two training interventions, while timing measurements quantify
their computational cost.

\begin{table*}[t]
\centering
\small
\caption{
    \textbf{Planning success ($\%$, $\uparrow$) across five continuous-control tasks.}
    Each entry reports mean $\pm$ sample standard deviation over six
    evaluation seeds.
    Both DINO-WM variants use a frozen pretrained encoder;
    $\dagger$ denotes additional proprioceptive state inputs, and
    $\ddagger$ denotes reachability-aware planning.
    \textbf{Bold} and \underline{underline} mark the highest and second-highest
    means per task, including ties; the shaded row is \name{}, and dashes
    mark settings not evaluated.
}
\label{tab:planning}
\setlength{\tabcolsep}{4pt}
\renewcommand{\arraystretch}{1.1}
\begin{tabularx}{\textwidth}{l*{5}{C}}
\toprule
& \multicolumn{3}{c}{\textbf{2D}} & \multicolumn{2}{c}{\textbf{3D}} \\
\cmidrule(lr){2-4}\cmidrule(lr){5-6}
\textbf{Method}
    & \textbf{Two-Room}
    & \textbf{Reacher}
    & \textbf{PushT}
    & \textbf{OGB-Cube}
    & \textbf{Scene} \\
\midrule
PLDM~\cite{sobal2025pldm}
    & \res{96.33}{2.94}
    & \res{78.67}{5.89}
    & \res{69.67}{6.74}
    & \res{58.00}{6.32}
    & \res{57.00}{3.95} \\
DINO-WM~\cite{zhou2025dino-wm}
    & \resu{99.00}{1.67}
    & \res{68.67}{4.32}
    & \res{53.33}{6.28}
    & \resb{73.33}{4.13}
    & \res{53.67}{4.97} \\
DINO-WM+proprio$^\dagger$~\cite{zhou2025dino-wm}
    & \resb{99.67}{0.82}
    & \nores
    & \res{80.00}{4.56}
    & \nores
    & \res{53.33}{6.02} \\
LeWM~\cite{maes_lelidec2026lewm}
    & \res{83.67}{4.80}
    & \resu{84.00}{5.66}
    & \res{83.67}{6.12}
    & \res{67.33}{5.01}
    & \resu{66.33}{4.63} \\
Fast-LeWM~\cite{gao2026fast-lewm}
    & \res{91.33}{3.93}
    & \res{81.67}{2.34}
    & \resu{88.00}{6.07}
    & \resu{73.00}{4.69}
    & \res{61.33}{4.50} \\
RC-aux~\cite{li2026predictive}
    & \res{90.33}{4.80}
    & \res{81.00}{7.46}
    & \res{77.00}{6.03}
    & \res{71.00}{5.33}
    & \nores \\
RC-aux+RA$^\ddagger$~\cite{li2026predictive}
    & \res{96.00}{3.10}
    & \res{86.00}{4.38}
    & \res{83.67}{5.28}
    & \nores
    & \nores \\
\midrule
\oursrow
\textbf{\name{} (Ours)}
    & \res{96.33}{4.46}
    & \resb{86.00}{4.00}
    & \resb{97.00}{2.45}
    & \res{69.67}{4.27}
    & \resb{71.00}{4.15} \\
\bottomrule
\end{tabularx}
\end{table*}

\subsection{Experimental Setup}
\label{subsec:setup}

\paragraph{Tasks and data.}
We evaluate \name{} on five continuous-control tasks spanning navigation,
reaching, and manipulation: Two-Room~\cite{zhou2025dino-wm},
Reacher~\cite{reacher}, and PushT~\cite{chi2025diffusion} in 2D, and
OGB-Cube and Scene~\cite{park2025ogbench} in 3D.
For each task, \name{} is trained on RGB observations and actions from
the source dataset.

\paragraph{Baselines.}
LeWM~\cite{maes_lelidec2026lewm} is our primary baseline for assessing
training changes, as \name{} retains its architecture and Euclidean-cost
planner. Within the LeWM family, Fast-LeWM~\cite{gao2026fast-lewm}
predicts future latents in parallel, while RC-aux~\cite{li2026predictive}
adds multi-horizon prediction and reachability supervision.
RC-aux+RA additionally modifies the planning cost using reachability
estimates. We also compare with PLDM~\cite{sobal2025pldm}, which uses
multiple auxiliary training objectives, and DINO-WM~\cite{zhou2025dino-wm},
which uses a frozen pretrained visual encoder. DINO-WM+proprio further
receives proprioceptive states as privileged inputs.

\paragraph{Training and evaluation protocol.}
The LeWM-family methods share a 50-epoch training budget.
PLDM and DINO-WM use officially released weights, without a matched
pretraining budget. LeWM-family comparisons use a planning horizon of
25 environment steps and a CEM budget of 300 candidates per iteration
for 30 iterations. We evaluate each trained model on 50 start--goal
episodes per evaluation seed, using the same six seeds across methods.
Each episode has a budget of 50 environment steps, with success
determined by the task's native criterion.
We report mean success rate $\pm$ sample standard deviation across
evaluation seeds.
Further implementation details are provided in the supplementary material.

\subsection{Planning Performance}
\label{subsec:planning}

\name{} achieves higher mean planning success than LeWM on all five
tasks in \cref{tab:planning}, while preserving LeWM's prediction and
planning interfaces.
The gains are larger on Two-Room and PushT, with smaller improvements
on Reacher, OGB-Cube, and Scene.
On PushT, mean success increases from $83.67\%$ to $97.00\%$, a gain of
$13.3$ percentage points.

\name{} has the highest mean success on PushT and Scene and ties
RC-aux+RA for the highest mean on Reacher in \cref{tab:planning}.
On Two-Room, its mean exceeds those of Fast-LeWM and RC-aux and is
close to that of RC-aux+RA.
OGB-Cube is the exception within the LeWM family, with higher mean
success for Fast-LeWM and RC-aux.
Pretrained DINO-WM leads on OGB-Cube, and both DINO-WM variants have
higher means than \name{} on Two-Room.

\begin{figure*}[t]
\centering
\includegraphics{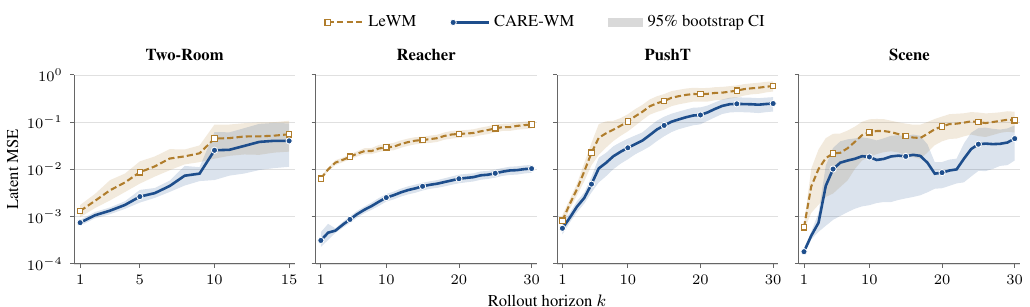}
\vspace{\horizonfigurecaptionspace}
\caption{
    \textbf{Open-loop rollout accuracy across prediction horizons.}
    Lines show mean latent MSE ($\downarrow$) over 100 paired trajectory
    windows per task, one per source-dataset episode; shaded bands show
    pointwise 95\% confidence intervals from 2,000 paired trajectory-bootstrap
    draws.
    Rollouts start from three observed frames and follow recorded actions,
    with five environment steps per rollout step.
    Latent targets come from each model's own encoder; all panels use a
    common logarithmic vertical scale.
}
\label{fig:horizon-gen}
\end{figure*}

\subsection{Rollout Accuracy \& Planning-Cost Alignment}
\label{subsec:predictive-fidelity}
\label{subsec:horizon-gen}
\label{subsec:planning-cost-fidelity}
\label{subsec:latent-cost-geometry}

To investigate the planning gains, we compare \name{} and LeWM
using the rollout accuracy and planning-cost alignment diagnostics
introduced in \cref{sec:background_planning}.
We evaluate recursive prediction error on paired trajectory windows,
using identical observed histories and recorded actions for both models.
For candidate rankings on Two-Room and PushT, we also include RC-aux;
all three models share initial states, goals, and candidate action sequences.
Sampling and execution details are given in
\cref{app:rollout-evaluation-protocol,app:latent-cost-geometry-protocol}
of the supplementary material.

\Cref{fig:horizon-gen} shows lower mean rollout MSE for \name{} than
for LeWM at every evaluated horizon on all four tasks.
The improvement is present at the planning horizon $k=5$, corresponding
to 25 environment steps, and extends to $k=15$ on Two-Room and $k=30$
on Reacher, PushT, and Scene.
Because the models use different encoders, we also divide each model's
MSE by the mean variance of its target latent coordinates.
The normalized error remains lower at the final evaluated horizon on
all four tasks.
This rules out uniform latent scale shrinkage as the sole explanation
for the reported reduction.

The rank correlations in \cref{tab:latent-cost-geometry} distinguish
preservation of latent-cost rankings from agreement with physical task costs.
Following \cref{sec:background_planning}, $\rho(\hat C,C)$ measures whether
prediction preserves the realized latent-cost ordering, while $\rho(C,J)$
measures whether that ordering agrees with physical task costs.
The correlation $\rho(\hat C,J)$ assesses the predicted ordering available
to the planner against those task costs.

\begin{table}[t]
\centering
\caption{
    \textbf{Alignment of candidate cost rankings.}
    Spearman correlations ($\uparrow$) use all 63 shared candidates at
    each of 2,048 common start--goal pairs per task and are averaged
    equally over starts.
    $\hat C$, $C$, and $J$ denote predicted latent goal, realized latent goal,
    and physical task costs, respectively.
    \textbf{Bold} marks the highest mean in each row.
}
\label{tab:latent-cost-geometry}
\begingroup
\footnotesize
\setlength{\tabcolsep}{3pt}
\renewcommand{\arraystretch}{1.1}
\begin{tabularx}{\linewidth}{llCCC}
\toprule
\textbf{Task} & \textbf{Pair} & \textbf{LeWM} & \textbf{RC-aux} & \textbf{\name{}} \\
\midrule
\multirow{3}{*}{PushT} & $\rho(C,J)$ & 0.9084 & 0.9066 & \textbf{0.9233} \\
 & $\rho(\hat C,C)$ & 0.8944 & 0.8895 & \textbf{0.9566} \\
 & $\rho(\hat C,J)$ & 0.8540 & 0.8489 & \textbf{0.9104} \\
\midrule
\multirow{3}{*}{Two-Room} & $\rho(C,J)$ & 0.8663 & 0.9204 & \textbf{0.9392} \\
 & $\rho(\hat C,C)$ & 0.9325 & 0.9712 & \textbf{0.9888} \\
 & $\rho(\hat C,J)$ & 0.8468 & 0.9036 & \textbf{0.9288} \\
\bottomrule
\end{tabularx}
\endgroup

\end{table}

\name{} achieves the highest correlations on all three cost pairs in both tasks.
On PushT, LeWM and RC-aux have similar correlations: realized latent costs
align closely with physical task costs, with $\rho(C,J)\approx0.91$,
but prediction weakens this alignment.
\name{} raises $\rho(\hat C,C)$ from LeWM's $0.8944$ to $0.9566$
and $\rho(\hat C,J)$ from $0.8540$ to $0.9104$.
Its rollout MSE at the planning horizon is also $78.3\%$ lower than LeWM's,
providing complementary evidence of more faithful prediction,
consistent with the planning gain.
On Two-Room, \name{} improves both ordering preservation and realized
latent-cost alignment over LeWM and RC-aux.

\subsection{Component Ablations}
\label{subsec:ablation}

\begin{table}[t]
\centering
\footnotesize
\caption{
\textbf{Component contributions to planning success ($\%$, $\uparrow$).}
SHSC and CA-SIGReg are added to LeWM individually or together as \name{}.
Each entry reports mean $\pm$ sample standard deviation over six evaluation
seeds for a fixed trained model, with 50 episodes per seed.
\textbf{Bold} marks the highest mean per task; the shaded row is \name{}.
}
\label{tab:ablation}
\setlength{\tabcolsep}{3pt}
\renewcommand{\arraystretch}{1.1}
\begin{tabularx}{\linewidth}{l*{4}{C}}
\toprule
\textbf{Method} & \textbf{Two-Room} & \textbf{Reacher} & \textbf{PushT} & \textbf{OGB-Cube} \\
\midrule
LeWM
    & \res{83.67}{4.80} & \res{84.00}{5.66} & \res{83.67}{6.12} & \res{67.33}{5.01} \\
\;+\,SHSC
    & \res{88.00}{7.16} & \res{85.67}{2.94} & \res{92.33}{4.27} & \res{68.33}{7.74} \\
\;+\,CA-SIGReg
    & \res{88.67}{5.16} & \resb{86.33}{5.13} & \res{91.67}{5.28} & \res{68.67}{7.34} \\
\midrule
\oursrow
\textbf{\name{}}
    & \resb{96.33}{4.46} & \res{86.00}{4.00} & \resb{97.00}{2.45} & \resb{69.67}{4.27} \\
\bottomrule
\end{tabularx}
\end{table}

SHSC and CA-SIGReg each improve mean planning success over LeWM on all
four tasks in \cref{tab:ablation}.
Their combination, \name{}, achieves the highest mean on Two-Room, PushT,
and OGB-Cube, while CA-SIGReg alone is slightly higher on Reacher.
We next examine SHSC's prediction behavior and CA-SIGReg's adaptive
weighting to understand these gains.

\subsection{Understanding SHSC}
\label{subsec:shsc-mechanism}
\label{subsec:mechanism}
\label{subsec:exposure-gap}

\paragraph{BPTT comparison.}
Our backpropagation through time (BPTT) reference replaces SHSC's single
student call with three recursive calls, extending the training window
from four to seven frames. Each recursive call reuses all three predictions
from the preceding call and receives direct supervision only on its last
prediction. The teacher loss and differentiable feedback are retained.
As shown in \cref{tab:feedback-ablation}, SHSC achieves similar mean planning
success and generated-context MSE on Two-Room, with higher success and lower
MSE on PushT. Implementation details are provided in the supplementary material.

\paragraph{Prediction under recursive feedback.}
To assess whether shallow feedback training improves deeper recursive
prediction, we compare LeWM and SHSC under observed and recursively
generated contexts at matched target times.
On PushT, LeWM and SHSC have nearly identical MSE with observed contexts,
$1.351$ and $1.322$ in $10^{-3}$ units.
After five recursive context updates, the MSEs for the next prediction are
$37.277$ and $25.134$, respectively, a $32.6\%$ reduction with SHSC.
The improvement mainly appears when predicted states are reused as inputs.
The full analysis is provided in \cref{app:context_diagnostics} of the
supplementary material.

\paragraph{Differentiable feedback ablation.}
SHSC (detached) retains training on generated inputs but blocks the
student gradient through those inputs.
It has lower mean planning success and higher generated-context MSE than
full SHSC on both tasks in \cref{tab:feedback-ablation}.
We further analyze the training dynamics of SHSC (detached) on PushT.
Between epochs 10 and 50, short-window validation losses decrease,
while mean five-step terminal error increases.
At epoch 50, refreshing the context with encoded observations at each step
substantially reduces the terminal error.
These observations support our motivation for differentiable feedback.
Both variants train the predictor to make predictions from generated latents,
while full SHSC additionally propagates subsequent prediction errors through these inputs,
directly guiding the generation of intermediate latents that better support later predictions.

\begin{table}[htbp]
\centering
\footnotesize
\caption{Planning success and generated-context prediction.
Success is the mean percentage over six evaluator seeds of 50 episodes.
MSE is in $10^{-3}$ units, averaged over $D=192$ coordinates and the
same 64 paired windows per environment. Starting from three observed
frames, $h=5$ recursive feedback updates precede the measured prediction
($k=6$). \textbf{Bold} marks the best value
per column; the shaded row is \name{}. The BPTT reference is
defined in \cref{subsec:shsc-mechanism}.}
\label{tab:feedback-ablation}
\setlength{\tabcolsep}{3pt}
\renewcommand{\arraystretch}{1.1}
\begin{tabularx}{\linewidth}{l*{4}{C}}
\toprule
& \multicolumn{2}{c}{\textbf{Two-Room}} & \multicolumn{2}{c}{\textbf{PushT}} \\
\cmidrule(lr){2-3}\cmidrule(lr){4-5}
\textbf{Method} & Success $\uparrow$ & MSE $\downarrow$
& Success $\uparrow$ & MSE $\downarrow$ \\
\midrule
LeWM & 83.67 & 8.606 & 83.67 & 37.277 \\
BPTT & 89.33 & \textbf{4.312} & 90.67 & 30.897 \\
SHSC (detached) & 85.67 & 11.051 & 52.67 & 31.839 \\
SHSC & 88.00 & 4.332 & 92.33 & 25.134 \\
\midrule
\oursrow
\textbf{\name{}} & \textbf{96.33} & 4.402 & \textbf{97.00} & \textbf{7.180} \\
\bottomrule
\end{tabularx}
\end{table}

\subsection{Understanding CA-SIGReg}
\label{subsec:ca-mechanism}

\begin{figure}[htbp]
\centering
\includegraphics{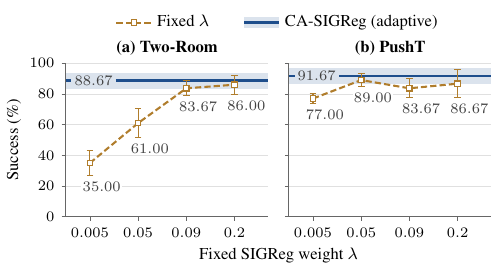}
\caption{
    \textbf{Planning success with adaptive and fixed SIGReg weights ($\%$, $\uparrow$).}
    All variants omit SHSC, with fixed-weight models using the LeWM objective.
    Fixed weights are equally spaced on the horizontal axis; markers show
    their mean success, and horizontal reference lines show the mean success
    of standalone CA-SIGReg.
    Means are computed over six evaluation seeds for each fixed trained
    model, with 50 episodes per seed; error bars and shaded bands show
    $\pm1$ sample standard deviation across seeds.
 }
\label{fig:ca-fixed-weights}
\end{figure}

\paragraph{Adaptive versus fixed SIGReg weights.}
We compare standalone CA-SIGReg with LeWM trained using fixed SIGReg
coefficients on Two-Room and PushT.
As shown in \cref{fig:ca-fixed-weights}, CA-SIGReg achieves the highest
mean planning success on both tasks.
On Two-Room, lowering the fixed coefficient from $0.09$ to $0.05$
reduces mean success from $83.67\%$ to $61.00\%$, showing that simply
reducing the coefficient does not consistently improve planning.

\paragraph{Relative regularization strength.}
Even with a fixed coefficient, the relative strength of prediction and
regularization gradients can change during training.
In Two-Room SHSC training with a fixed SIGReg coefficient of $0.09$, the
median prediction-gradient norm on the observation encoder decreases from
$0.509$ in epochs 1--5 to $0.150$ in epochs 26--50.
Over the same periods, the median weighted SIGReg gradient norm decreases
from $0.508$ to $0.210$.
Regularization gradients therefore weaken more slowly and become stronger
relative to prediction gradients.
This changing balance motivates adapting the coefficient during training.
CA-SIGReg reduces it when the calibrated discrepancy from the Gaussian
target is small.
Measurement details are provided in \cref{app:persistent-pressure} of the
supplementary material.

\begin{figure}[htbp]
\centering
\includegraphics{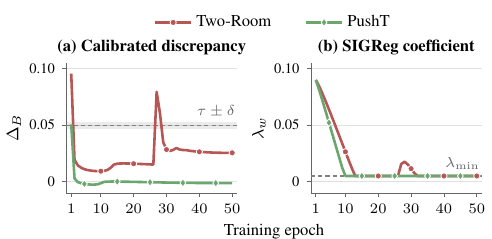}
\caption{
    \textbf{CA-SIGReg adaptation during \name{} training.}
    For one run per task, (a) shows the epoch-mean calibrated discrepancy
    and (b) the epoch-end coefficient.
    Shading marks the tolerance band; dashed lines mark its center and
    the coefficient floor.
    Coefficient updates use the discrepancy EMA updated at each
    optimizer step.
}
\label{fig:training-dynamics}
\end{figure}

\paragraph{Adaptive training dynamics.}
\Cref{fig:training-dynamics} shows one full \name{} training run for
Two-Room and one for PushT.
On both tasks, the coefficient decreases from $0.09$ to its lower bound
of $0.005$ as the calibrated discrepancy falls below the tolerance band.
On Two-Room, the discrepancy rises above the band at epochs 27--28,
accompanied by a temporary increase in the coefficient from $0.005$ to
approximately $0.018$.
The coefficient subsequently decreases again, illustrating that the
controller both relaxes regularization and strengthens it when the
discrepancy grows.

\subsection{Computational Efficiency}
\label{subsec:training-cost}

The measurements cover complete training updates, including encoding all
frames required by each method.
As shown in \cref{tab:training-cost}, \name{} takes approximately
$15\%$ more time per update than LeWM on PushT and Two-Room, with the
same reported peak allocated memory of $12.44$\,GiB.
Compared with the BPTT reference evaluated in \cref{subsec:shsc-mechanism},
\name{} takes about $40\%$ less time per update and uses $42\%$ less peak allocated memory.
The measurement protocol is provided in \cref{app:training-cost-protocol}
of the supplementary material.

\begin{table}[htbp]
\centering
\footnotesize
\caption{
    \textbf{Per-update training time and peak allocated memory ($\downarrow$).}
    Measurements use a single RTX 4090, batch size 128, and BF16; data
    loading, transfer, logging, and epoch-end diagnostics are excluded.
    Time is mean $\pm$ sample standard deviation of the mean update times
    from three repetitions, each with 20 warmup and 50 measured updates;
    memory is the maximum peak allocated memory across the three repetitions.
}
\label{tab:training-cost}
\setlength{\tabcolsep}{3pt}
\renewcommand{\arraystretch}{1.1}
\begin{tabularx}{\linewidth}{lCcCc}
\toprule
& \multicolumn{2}{c}{\textbf{PushT}} & \multicolumn{2}{c}{\textbf{Two-Room}} \\
\cmidrule(lr){2-3}\cmidrule(lr){4-5}
\textbf{Method} & ms/update & GiB & ms/update & GiB \\
\midrule
LeWM      & \res{153.68}{0.03} & 12.44 & \res{154.86}{0.10} & 12.44 \\
SHSC      & \res{174.44}{0.18} & 12.44 & \res{175.54}{0.06} & 12.44 \\
CA-SIGReg & \res{168.43}{0.13} & 12.44 & \res{169.41}{0.52} & 12.44 \\
\oursrow
\textbf{\name{}} & \res{176.85}{0.30} & 12.44 & \res{177.37}{0.25} & 12.44 \\
\midrule
BPTT & \res{296.79}{0.28} & 21.63 & \res{298.09}{0.32} & 21.63 \\
\bottomrule
\end{tabularx}

\end{table}

\begin{conclusionblock}
\section*{Conclusion}\label{sec:conclusion}

\name{} improves planning reliability through training alone, retaining
LeWM's architecture and Euclidean-cost planner.
SHSC trains the predictor using its own predictions as inputs within the
original supervision window, while CA-SIGReg adapts the regularization
weight using a finite-sample-calibrated discrepancy.
Across five tasks, \name{} improves mean planning success over LeWM and
matches or exceeds the highest mean among other LeWM-family methods on four.
Diagnostics show lower rollout error, better preservation of realized
latent-cost rankings, and closer alignment between realized costs and
physical task costs.
With modest per-update overhead over LeWM and lower time and memory costs
than our BPTT reference, \name{} offers a practical route to more reliable
latent-space planning.
\end{conclusionblock}

\ifdefined\carearxiv\else
  \clearpage
\fi
\section*{LLM Usage Statement}
We used AI tools in preparing this work.
Tool-generated manuscript text, code and implementations, and figures and
diagrams were reviewed, edited, or approved by the authors. Assistance with
related work and citations, experimental design, execution and analysis,
and research ideas, hypotheses, claims and conclusions was limited to
editing or refining author-produced content. The full manuscript or major
portions were also shared with OpenAI (ChatGPT/GPT) and Anthropic (Claude)
for reviewer-style feedback.
No LLMs or other AI tools were used for synthetic data generation, dataset
creation, or annotation. The authors take full responsibility for all
submitted content.

{
  \small
  \bibliographystyle{ieeenat_fullname}
  \bibliography{bib/eg,bib/exp_bias,bib/bptt,bib/covariate_shift,bib/misc}
}

\clearpage
\ifdefined\carearxiv\else
  \setcounter{page}{1}
\fi
\maketitlesupplementary
\appendix
\section{SIGReg Calibration and Scale Gradients}
\label{app:theory}

This appendix defines the statistic used in \cref{sec:ca_reg}, derives its
Gaussian and zero-collapse references, and examines its scale gradient at
the Gaussian target. The generated-context error decomposition is given
with the SHSC diagnostics in \cref{app:context_decomposition}.

\subsection{SIGReg Statistic}
\label{app:sigreg_definition}

LeWM uses Sketched Isotropic Gaussian Regularization
(SIGReg)~\cite{balestriero2025lejepa,maes_lelidec2026lewm} to compare
projected encodings with a standard normal reference. Let
$Z=(z_{s,b})\in\mathbb R^{T\times B\times D}$ collect a minibatch of $B$
encoded windows, with $T=K+1$ time positions. Draw $M$ random unit
directions $u_m\in\mathbb R^D$. At each position $s$, each comparison uses
the $B$ projections $\langle z_{s,b},u_m\rangle$; samples from different
time positions are not pooled.

At frequency node $t_k$, the empirical characteristic function and its
standard Gaussian reference are
\begin{equation}
    \begin{aligned}
        \widehat\varphi_{s,m}(t_k)
        &=\frac{1}{B}\sum_{b=1}^{B}
          e^{\mathrm{i}t_k\langle z_{s,b},u_m\rangle},\\
        \varphi(t_k)&=e^{-t_k^2/2},
    \end{aligned}
    \label{eq:sigreg_ecf}
\end{equation}
where $\mathrm{i}=\sqrt{-1}$ and $t_k$ denotes frequency, not environment
time. The discrete Epps--Pulley discrepancy is
\begin{equation}
    \mathrm{EP}_{s,m}
    =B\sum_k\omega_k
    \left|\widehat\varphi_{s,m}(t_k)-\varphi(t_k)\right|^2,
    \label{eq:sigreg_ep}
\end{equation}
where each nonnegative $\omega_k$ combines the Gaussian window and
trapezoidal quadrature weight~\cite{balestriero2025lejepa,epps-pulley}.
The raw penalty retains the factor $B$ and averages over positions and
projections:
\begin{equation}
    R(Z)=\mathrm{SIGReg}(Z)
    =\frac{1}{TM}\sum_{s=1}^{T}\sum_{m=1}^{M}\mathrm{EP}_{s,m}.
    \label{eq:sigreg_statistic}
\end{equation}
This finite-sample discrepancy responds to scale, anisotropy, rank loss,
and distribution-shape mismatch. It does not specify which task-relevant
distances the representation preserves.

\subsection{Finite-Sample Calibration}
\label{app:sigreg_references}

The references in Eq.~\eqref{eq:ca_health} use the same fixed quadrature
nodes and weights as Eq.~\eqref{eq:sigreg_ep}. Projection directions have
unit norm and are sampled independently of the embeddings. Gaussian
expectations below are conditional on these directions.

\paragraph{Gaussian reference.}
At each position $s$, assume that $z_{s,1},\ldots,z_{s,B}$ are independent
$\mathcal N(0,I_D)$ samples. For a fixed projection $m$, the variables
$x_b=\langle z_{s,b},u_m\rangle$ are then independent standard normals.
Since $\mathbb E[e^{\mathrm{i}t_kx_b}]=\varphi(t_k)$ and
$|e^{\mathrm{i}t_kx_b}|=1$, the empirical characteristic function is
unbiased and its mean squared error is
\begin{equation}
    \mathbb E\left[
        |\widehat\varphi_{s,m}(t_k)-\varphi(t_k)|^2\right]
    =\frac{1-\varphi(t_k)^2}{B}.
    \label{eq:app_ecf_error}
\end{equation}
Substituting into Eq.~\eqref{eq:sigreg_ep} cancels the factor $1/B$.
Averaging over positions and projections therefore gives
\begin{equation}
    R_{\mathrm N}=\mathbb E[R(Z)]
    =\sum_k\omega_k\bigl(1-e^{-t_k^2}\bigr).
    \label{eq:app_gaussian_reference}
\end{equation}
Under this sampling model, $R_{\mathrm N}$ is independent of $B$.
By linearity of expectation, this expression holds even when the
statistics $\mathrm{EP}_{s,m}$ are dependent across positions or projections.

\paragraph{Zero-collapse reference.}
When every embedding is zero, every empirical characteristic function is
one. Each position and projection then has the same statistic, yielding
\begin{equation}
    R_{\mathrm C}(B)=R(0)
    =B\sum_k\omega_k\bigl(1-e^{-t_k^2/2}\bigr)^2.
    \label{eq:app_collapse_reference}
\end{equation}
This deterministic reference grows linearly with $B$. It is also a lower
bound for any point collapse at each time position. To see this, let all
embeddings at position $s$ equal $\bar z_s$ and set
$x=\langle\bar z_s,u_m\rangle$. Then
\begin{equation}
    \begin{aligned}
        |e^{\mathrm{i}t_kx}-\varphi(t_k)|^2
        &=1-2\varphi(t_k)\cos(t_kx)+\varphi(t_k)^2\\
        &\geq (1-\varphi(t_k))^2,
    \end{aligned}
    \label{eq:app_point_collapse}
\end{equation}
since $\varphi(t_k)>0$ and $\cos(t_kx)\leq1$. Nonnegative quadrature
weights preserve this inequality, so $R(Z)\geq R_{\mathrm C}(B)$ for
such collapses.

The calibrated discrepancy $\Delta_B(Z)$ in Eq.~\eqref{eq:ca_health}
has expectation zero under the Gaussian sampling model and equals one
when every encoding is zero. At the training batch size $B=128$,
$R_{\mathrm C}(128)>R_{\mathrm N}$, so $\Delta_B(Z)\geq1$ whenever the
$B$ encodings coincide at each time position, even if the common encoding
varies across positions.

\subsection{Scale Gradients at the Gaussian Target}
\label{app:sigreg_scale_gradient}

A nonzero reference value alone does not establish a nonzero gradient.
To examine the gradient, consider one unit projection with
$x_b=\sigma\epsilon_b$, where $\epsilon_b$ are independent standard normal
variables and $\sigma>0$. Write
$\varphi_\sigma(t)=e^{-\sigma^2t^2/2}$, and let $R_B(\sigma)$ denote the
batch-scaled EP statistic for this projection. The empirical characteristic
function of $x_1,\ldots,x_B$ has mean $\varphi_\sigma$ and variance
$(1-\varphi_\sigma^2)/B$, so
\begin{equation}
    \begin{aligned}
        \mathbb E R_B(\sigma)
        &=\sum_k\omega_k\bigl[1-\varphi_\sigma(t_k)^2\bigr]\\
        &\quad+B\sum_k\omega_k
            \bigl[\varphi_\sigma(t_k)-\varphi(t_k)\bigr]^2.
    \end{aligned}
    \label{eq:app_scale_expectation}
\end{equation}
The first term comes from sampling variance. The second is $B$ times the
population discrepancy on the quadrature grid. Since $\varphi_1=\varphi$,
each squared difference has zero derivative at $\sigma=1$. The sampling term gives
\begin{equation}
    \left.\frac{\mathrm d}{\mathrm d\sigma}\mathbb E R_B(\sigma)
    \right|_{\sigma=1}
    =2\sum_k\omega_k t_k^2e^{-t_k^2}>0.
    \label{eq:app_scale_gradient}
\end{equation}
The derivative is positive because the quadrature assigns positive weight
to nonzero frequencies. A sufficiently small decrease from $\sigma=1$
therefore lowers the expected penalty. For a Gaussian batch $Z$ as in
\cref{app:sigreg_references}, linearity of expectation gives the same
derivative for $\mathbb E[R(\sigma Z)]$ after averaging over positions and
unit projections.

The slope of $\mathbb E[R_B(\sigma)/B]$ at $\sigma=1$ is proportional
to $1/B$, consistent with the finite-batch gradient bias LeJEPA derives
for the unscaled discrepancy~\cite{balestriero2025lejepa}. The factor $B$
in Eq.~\eqref{eq:sigreg_ep} makes the slope of the expected raw penalty
at the target independent of batch size. Subtracting the constant
$R_{\mathrm N}$ preserves this slope. CA-SIGReg uses a detached moving
average of $\Delta_B(Z)$ to adjust $\lambda_w$, which scales the gradient
of $R(Z)$ within each update window. The Gaussian probe in
\cref{app:persistent-pressure} checks the derivative in
Eq.~\eqref{eq:app_scale_gradient}.

\section{Experimental Details}
\label{app:experimental_details}

We describe the architecture and training procedure for \name{} on the
benchmarks introduced in \cref{subsec:setup}. We then give the protocols
for planning evaluation, model diagnostics, and training-cost measurements.

\subsection{Model Architecture}
\label{app:model-architecture}

The encoder and predictor in \cref{tab:model-architecture} follow the
LeWM architecture~\cite{maes_lelidec2026lewm}. A ViT-Tiny/14 trained from
scratch processes each $224\times224$ RGB image as 256 patch tokens and one
CLS token. The encoder projector maps the final CLS token to a
192-dimensional latent. The causal Transformer
predictor takes a sequence of these latents, adds learned temporal
position embeddings, and conditions each block on the actions through
AdaLN-Zero. A separate prediction projector maps each output back to the
encoder's latent space.

Each model step groups five consecutive, separately specified environment
actions. A pointwise 1D convolution maps their $5d_a$ concatenated
coordinates to 10 channels, followed by a $10\to768\to192$ MLP with SiLU.
The action dimension $d_a$ is 2 for Two-Room, Reacher, and PushT, and 5
for OGB-Cube and Scene. Images and actions are the model's only inputs.

\begin{table}[htbp]
\centering
\footnotesize
\caption{
    \textbf{Model architecture of \name{} on all five tasks.}
    In each predictor block, an output projection maps the 1,024
    concatenated attention features ($16$ heads of width $64$) back to
    the model width of 192. The encoder and prediction projectors have
    separate parameters and apply BatchNorm and GELU between their linear
    layers. The action encoder is a pointwise convolution ($5d_a\to10$)
    followed by an MLP with SiLU.
}
\label{tab:model-architecture}
\setlength{\tabcolsep}{4pt}
\renewcommand{\arraystretch}{1.1}
\begin{tabularx}{\linewidth}{l>{\hsize=0.9\hsize}C>{\hsize=1.1\hsize}C}
\toprule
\textbf{Setting} & \textbf{Encoder} & \textbf{Predictor} \\
\midrule
Architecture & ViT-Tiny/14 & Causal Transformer \\
Input & $224\times224$ RGB & Projected latents \\
Patch size & $14\times14$ & \nores \\
Position embeddings & Learned spatial & Learned temporal \\
Max.\ temporal context & \nores & 3 model steps \\
\addlinespace
Transformer blocks & 12 & 6 \\
Hidden width & 192 & 192 \\
Attention heads & 3 & 16 \\
Dimension per head & 64 & 64 \\
Feed-forward width & 768 & 2048 \\
Feed-forward activation & GELU & GELU \\
Attention dropout & 0 & 0.1 \\
MLP dropout & 0 & 0.1 \\
Embedding dropout & 0 & 0 \\
\addlinespace
Action conditioning & \nores & AdaLN-Zero \\
Action encoder & \nores & $5d_a{\to}10{\to}768{\to}192$ \\
Output latents & CLS token & One per input step \\
Output projector & $192{\to}2048{\to}192$ & $192{\to}2048{\to}192$ \\
\bottomrule
\end{tabularx}
\end{table}

\subsection{Training Configuration}
\label{app:training-configuration}

We scale RGB values to $[0,1]$, normalize the channels using the ImageNet
means $(0.485,0.456,0.406)$ and standard deviations $(0.229,0.224,0.225)$,
and resize the images to $224\times224$. We standardize each action
coordinate using its mean and standard deviation in the source dataset.

We follow LeWM's protocol for splitting the data into training and
validation sets and forming training batches~\cite{maes_lelidec2026lewm}.
For planning, we select LeWM-family checkpoints by validation loss
within the shared training budget in \cref{tab:training-configuration}.
We evaluate PLDM and DINO-WM using their officially released weights.

\begin{table}[tbp]
\centering
\footnotesize
\caption{
    \textbf{Training configuration of \name{}.}
    With zero-based epochs $e$, only the teacher contributes to the SHSC
    prediction loss at $e=0$. The teacher and student losses have equal
    weight from $e=25$ onward.
}
\label{tab:training-configuration}
\setlength{\tabcolsep}{4pt}
\renewcommand{\arraystretch}{1.1}
\begin{tabularx}{\linewidth}{lL}
\toprule
\textbf{Setting} & \textbf{Value} \\
\midrule
\multicolumn{2}{l}{\textit{Data}} \\
Training window & 4 observations \\
Teacher context & 3 observations \\
Observation stride & 5 environment steps \\
Action block & 5 environment steps \\
Training / validation split & 90\% / 10\%, by window \\
Split and training shuffle seed & 3072 \\
Batch size per device & 128 for training and calibration \\
\midrule
\multicolumn{2}{l}{\textit{Optimization}} \\
Training budget & 50 epochs \\
Optimizer & AdamW \\
Peak learning rate & $5\times10^{-5}$ \\
Initial / minimum learning rate & 0 / 0 \\
Learning-rate schedule & Linear warmup, then cosine decay \\
Learning-rate warmup & 1\% of estimated optimizer updates, rounded down \\
Weight decay & $10^{-3}$ \\
Gradient-norm cap & 1.0 \\
Training precision & BF16 mixed precision \\
\midrule
\multicolumn{2}{l}{\textit{Same-Horizon Self-Conditioning}} \\
Student weight $\alpha_e$ & $\min(1,2e/50)$ \\
\midrule
\multicolumn{2}{l}{\textit{Constraint-Adaptive SIGReg}} \\
Random unit projections & 1,024 \\
Quadrature nodes / interval & 17 / $[0,3]$ \\
EMA decay $\beta$ & 0.95 \\
Initial coefficient $\lambda_0$ & 0.09 \\
Coefficient bounds $[\lambda_{\min},\lambda_{\max}]$ & $[0.005,0.5]$ \\
Epoch-level gain $\eta_\lambda$ & 0.20 \\
Discrepancy tolerance $\tau$ & 0.05 \\
Deadband half-width $\delta$ & 0.003 \\
Update interval $U$ & 10 successful optimizer steps \\
Controller warmup $E_{\mathrm w}$ & 1 epoch with EMA active \\
\bottomrule
\end{tabularx}
\end{table}

\subsection{Training Algorithm}
\label{app:training-algorithm}

In Algorithm~\ref{alg:care}, each predictor call processes its time positions
in parallel, and the student loss backpropagates through the teacher
predictions it receives in line~\ref{alg:care:student}. Each microbatch
contributes the penalty $\lambda_w R(Z)$ to the training loss and the
detached, calibrated discrepancy $\Delta_B(Z)$ to the controller
(lines~\ref{alg:care:loss}--\ref{alg:care:accumulate}).

At successful optimizer step $s$, $\Delta_s$ is the sum of these discrepancies
divided by the total number of microbatches across devices.
We discard feedback from skipped attempts and carry the exponential moving
average (EMA) across epoch boundaries, including the end of warmup. For
decay $0\leq\beta<1$, the update is
\begin{equation}
    \bar\Delta_s =
    \begin{cases}
        \Delta_1, & s=1,\\
        \beta\bar\Delta_{s-1}+(1-\beta)\Delta_s, & s\geq2.
    \end{cases}
    \label{eq:ca_ema}
\end{equation}

The operators in the coefficient update of Eq.~\eqref{eq:ca_controller} are
\begin{equation}
\begin{aligned}
    \operatorname{deadband}_{\delta}(x)
       &=\operatorname{sign}(x)\max\!\bigl(|x|-\delta,0\bigr),\\
    \operatorname{clip}(x)
       &=\min\!\bigl(\lambda_{\max},\max(\lambda_{\min},x)\bigr).
\end{aligned}
\label{eq:ca_operators}
\end{equation}

All devices share $\lambda_w$ throughout update window $w$. For a window
of $m_w$ successful optimizer steps, the effective step size is
\begin{equation}
    \eta_w=\eta_\lambda\frac{m_w}{N_e},
    \label{eq:ca_effective_step_size}
\end{equation}
where $N_e$ is the number of optimizer steps planned for the full epoch
given the local batch count and gradient accumulation. If steps are
skipped or the epoch ends early, $N_e$ stays fixed and the final window
uses its actual step count. Fresh runs start at $s=w=0$. On resumption,
we restore the coefficient, EMA, and window state from the checkpoint.

\begin{algorithm}[tbp]
\caption{\name{} training}
\label{alg:care}
\small
\renewcommand{\algorithmicrequire}{\textbf{Input:}}
\renewcommand{\algorithmicensure}{\textbf{Output:}}
\renewcommand{\algorithmiccomment}[1]{\hfill\(\triangleright\)~#1}
\begin{algorithmic}[1]
\REQUIRE windows $(o_{t:t+K},a_{t:t+K-1})$; settings in \cref{tab:training-configuration}
\ENSURE encoder $f_\psi$ and predictor $\pred_\phi$
\STATE initialize $(\psi,\phi)$; $s\gets0$; $w\gets0$
\FOR{epoch $e=0,\ldots,49$}
    \STATE $\alpha_e\gets\min(1,2e/50)$; $m_w\gets0$; $N_e\gets$ planned steps
    \FOR{each optimizer attempt with microbatches $\mathcal G$}
        \FOR{each microbatch $(o_{t:t+K},a_{t:t+K-1})\in\mathcal G$}
            \STATE $Z\gets f_\psi(o_{t:t+K})$%
                \COMMENT{$Z=z_{t:t+K}$}%
            \STATE $\hat Z^T\gets\pred_\phi(z_{t:t+K-1},a_{t:t+K-1})$%
                \COMMENT{\eqref{eq:teacher_path}}%
            \STATE\label{alg:care:student}%
                $\hat Z^S\gets\pred_\phi(\hat z^T_{t+1:t+K-1},a_{t+1:t+K-1})$%
                \COMMENT{\eqref{eq:student_path}}%
            \STATE\label{alg:care:loss}%
                $\mathcal L\gets(\mathcal L_T+\alpha_e\mathcal L_S)/(1+\alpha_e)
                +\lambda_w R(Z)$%
                \COMMENT{\eqref{eq:total_loss}}%
            \STATE\label{alg:care:accumulate}%
                accumulate $\nabla_{\psi,\phi}\mathcal L$ and detached $\Delta_B(Z)$%
                \COMMENT{\eqref{eq:ca_health}}%
        \ENDFOR
        \STATE attempt a clipped AdamW update of $(\psi,\phi)$
        \IF{the optimizer step succeeds}
            \STATE $s\gets s+1$; $\Delta_s\gets$ mean $\Delta_B(Z)$ over
                $\mathcal G$ and devices
            \STATE $\bar\Delta_s\gets\beta\bar\Delta_{s-1}+(1-\beta)\Delta_s$%
                \COMMENT{\eqref{eq:ca_ema}}%
            \STATE \textbf{if} $e\geq E_{\mathrm w}$ \textbf{then} $m_w\gets m_w+1$
        \ENDIF
        \IF{$m_w=U$ \OR ($m_w>0$ \AND epoch $e$ ends)}
            \STATE $\eta_w\gets\eta_\lambda m_w/N_e$%
                \COMMENT{\eqref{eq:ca_effective_step_size}}%
            \STATE $\lambda_{w+1}\gets\operatorname{clip}\bigl(\lambda_w
                +\eta_w\operatorname{deadband}_\delta(\bar\Delta_s-\tau)\bigr)$%
                \COMMENT{\eqref{eq:ca_controller}}%
            \STATE $w\gets w+1$; $m_w\gets0$
        \ENDIF
    \ENDFOR
\ENDFOR
\end{algorithmic}
\end{algorithm}

\subsection{BPTT Reference}
\label{app:bptt-reference}

Both the SHSC comparison and the training-cost measurements use the same
BPTT reference. It encodes seven frames $o_{t:t+6}$ once and makes four
sequential calls to the shared predictor. The teacher predicts the latents
at $t+1:t+3$ from observed latents $z_{t:t+2}$ and actions $a_{t:t+2}$.
Each of the three recursive calls takes all three
outputs of the preceding call as its context and shifts the action
sequence forward by one step. Their outputs therefore span $t+2:t+4$,
$t+3:t+5$, and $t+4:t+6$, respectively.

The teacher loss $\mathcal L_T$ supervises all three teacher predictions.
Each recursive call receives direct supervision only on its last prediction,
with targets $z_{t+4}$, $z_{t+5}$, and $z_{t+6}$ in order. Later
losses backpropagate through the generated inputs into earlier calls,
and the target latents also carry gradients to the encoder. We train the
encoder and predictor with these prediction losses and a fixed-weight SIGReg
penalty $0.09\,R(Z)$ on the encodings of all seven frames.

\subsection{Planning and Evaluation Configuration}
\label{app:planning-configuration}

For each start sampled from the task dataset, the goal state and image
come from the same trajectory 25 environment steps later. We use simulator
states to initialize episodes and evaluate the task's native success
criterion. Across evaluation seeds, we keep each model's parameters and
normalization statistics fixed and disable dropout.

\name{} plans from images and candidate actions. Its cost sums the
squared differences between the predicted terminal latent and the
encoded goal, giving 192 times the cost in Eq.~\eqref{eq:cem_cost}
and the same candidate ranking.

All LeWM-family planners share the horizon, candidate count, and number
of CEM iterations in \cref{tab:planning-configuration}.
LeWM and RC-aux roll out five model steps.
Fast-LeWM predicts the terminal latent directly from the full action
prefix. Both RC-aux variants train with
reachability supervision. \Cref{tab:planning} reports latent-distance
scoring for RC-aux on Two-Room, Reacher, PushT, and OGB-Cube. We evaluate
RC-aux+RA only on the first three tasks, with reachability added to the
planning cost.

\begin{table}[htbp]
\centering
\footnotesize
\caption{
    \textbf{Planning and evaluation settings of \name{} on all five tasks.}
    Each CEM candidate specifies a separate action for each of the
    25 environment steps in the planning horizon.
}
\label{tab:planning-configuration}
\setlength{\tabcolsep}{4pt}
\renewcommand{\arraystretch}{1.1}
\begin{tabularx}{\linewidth}{lL}
\toprule
\textbf{Setting} & \textbf{Value} \\
\midrule
\multicolumn{2}{l}{\textit{Latent CEM planner}} \\
Candidates per iteration & 300 \\
Iterations & 30 \\
Elite candidates & 30 \\
Initial Gaussian sampling scale & 1.0 in standardized action space \\
\midrule
\multicolumn{2}{l}{\textit{Horizon and execution}} \\
Planning horizon & 5 model steps (25 environment steps) \\
Actions per model step & 5 consecutive controls \\
Execution before replanning & 25 environment steps \\
Goal offset from start & 25 environment steps \\
Episode budget & 50 environment steps \\
\midrule
\multicolumn{2}{l}{\textit{Evaluation}} \\
Start--goal episodes per seed & 50 \\
Evaluation seeds & 42, 100, 2026, 3407, 1234, 4444 \\
Reported success rate & Mean $\pm$ sample SD across seeds \\
\bottomrule
\end{tabularx}
\end{table}

\subsection{Open-Loop Rollout Evaluation}
\label{app:rollout-evaluation-protocol}

On Two-Room, Reacher, PushT, and Scene, we sample one window from each
of 100 distinct source-dataset episodes per task. \name{} and LeWM encode
the same three initial frames and feed back their predictions under the
recorded actions. Each model step spans five environment steps. Rollouts
extend to $k=15$ on Two-Room and $k=30$ on the other three tasks.

At horizon $k$, we average $\|\hat z_k-z_k\|_2^2/192$ over windows using
each model's own encoding $z_k$ of the target frame.
For each model and task, we pool target encodings
across windows and horizons. The scale control divides MSE by the mean
of the 192 coordinate-wise variances.

\subsection{Generated-Context Diagnostic Protocols}
\label{app:context_probe_protocol}

On Two-Room and PushT, LeWM, SHSC, and \name{} share 64 windows per
task for the probes in \cref{app:context_diagnostics}. We sample one
valid start uniformly from each of 64 distinct source-dataset episodes and
take nine frames spaced five environment steps apart. We hold each
checkpoint fixed and reuse LeWM's open-loop evaluation checkpoints.

\paragraph{Recursive generated contexts.}
We initialize each rollout with three observed frames. Each update drops
the oldest latent and appends the next prediction under the recorded
actions. After $h\in\{1,2,5\}$ updates, we score prediction $k=h+1$.
The observed control encodes the frames at the same three context
positions and uses the same actions and target.

\paragraph{Two-call generated contexts.}
The causal teacher call predicts from $z_{t:t+2}$ and $a_{t:t+2}$. As in
SHSC and \name{} training, the student takes the first two teacher predictions
and the shifted actions $a_{t+1:t+2}$ to predict $z_{t+2}$ and $z_{t+3}$.

\paragraph{Effect of shortening the observed history.}
At the same two target positions, we compare the full teacher call with
a call that omits $z_t$. Replacing its two observed inputs $z_{t+1:t+2}$
with the teacher predictions then tests generated inputs at fixed history
length and actions. \Cref{tab:context-two-call} reports
the mean MSE over both targets ($S_{\mathrm{all}}$) and the MSE at
$t+3$ ($S_{\mathrm{last}}$).

\subsection{Gaussian Scale Probe Protocol}
\label{app:sigreg_probe_protocol}

For each Gaussian batch $Z$, we use automatic differentiation to measure
the scale derivative of the raw SIGReg statistic $R(\sigma Z)$ at
$\sigma=1$. Each repeat holds the standard Gaussian samples ($B=128$,
$D=192$) and 128 unit projection directions fixed across scales. Training
uses 1,024 directions. We run the probe in FP64 after casting the 17
quadrature nodes and weights from FP32. \Cref{app:persistent-pressure}
compares the mean derivative over 64 independent repeats with
Eq.~\eqref{eq:app_scale_gradient} and reports its standard error.

\subsection{Candidate-Ranking Evaluation Protocol}
\label{app:latent-cost-geometry-protocol}

\paragraph{Candidate construction.}
On Two-Room and PushT, we sample 2,048 starts per task from distinct
episodes with sufficient observation history and 25 subsequent actions.
Executing the unperturbed demonstration for 25 steps defines the goal
state and image. At each start, we generate 21 candidates at each noise
standard deviation ($0.1$, $0.3$, and $1.0$). We standardize actions
using the dataset's coordinate-wise means and standard deviations. For
each candidate, we draw one independent two-dimensional Gaussian offset
per five-step block and add it to all five demonstration actions in that
block. We convert back to simulator units and clip to $[-1,1]$, yielding
63 distinct perturbed sequences. We use seed 42 to sample the starts and
perturbations before candidate scoring or execution.

\paragraph{Execution and latent scoring.}
Before each candidate at a given start, we reset the simulator with the
same state and seed and replay the same action prefix. This restores the
start state and three observed frames spaced five environment steps apart.
We execute all 25 actions, even after early success, and apply the task's
native success criterion at the terminal state. Each model computes
$\hat C_i$ by rolling out recursively from these frames under the candidate
actions and $C_i$ by encoding the executed terminal image. Both costs
average squared coordinate differences from the model's goal encoding
directly in the projected latent space
(Eqs.~\eqref{eq:cem_cost} and~\eqref{eq:diagnostic_costs}).

\paragraph{Task geometry costs.}
On Two-Room, the physical cost $J_i$ is the shortest path length from
the executed terminal position to the goal on an eight-neighbor grid
with one-pixel spacing. Paths respect the simulator's collision
boundaries and exclude diagonal corner cuts. On PushT, we use
\begin{equation}
    J_i=\max\!\left(\frac{d_{\mathrm{pos}}}{20},\,
                    \frac{d_{\mathrm{angle}}}{\pi/9}\right),
\end{equation}
where $d_{\mathrm{pos}}$ is the Euclidean distance between the concatenated
agent and block positions at the terminal and goal states.
The angular error $d_{\mathrm{angle}}$ is the smallest absolute
difference between the two block orientations modulo $2\pi$.

\paragraph{Rank correlations.}
For \name{}, LeWM, and RC-aux, we compute the Spearman correlations
$\rho(\hat C,C)$, $\rho(C,J)$, and $\rho(\hat C,J)$ across all 63 shared
candidates at each start. \Cref{tab:latent-cost-geometry} reports each
correlation averaged equally over the 2,048 starts per task.

\paragraph{Success--failure inversions.}
For $k\in\{63,16,4\}$, LeWM and \name{} each select the $k$ candidates
with the lowest predicted costs under that model. A start is
eligible when each shortlist contains both successes and failures.
Within each model's shortlist, the inversion rate is the fraction of
success--failure pairs in which the failed candidate has strictly lower
predicted cost than the successful one. We average these per-start rates
equally over the common eligible starts for that $k$.
\Cref{fig:latent-cost-inversions} gives the counts by task and shortlist size.

\subsection{Training Cost Protocol}
\label{app:training-cost-protocol}

\Cref{tab:training-cost} reports time per update and peak allocated GPU
memory on PushT and Two-Room. Within each task, all methods use the same
dedicated NVIDIA GeForce RTX 4090, BF16 mixed precision, and batches of
128 windows. All use the LeWM architecture with a three-frame initial
context. Per four-frame window, LeWM and CA-SIGReg make one predictor
call, and SHSC and \name{} make two differentiable calls.

The methods share a fresh initialization, sampled trajectory starts,
action normalization, and AdamW settings. Each repetition reuses one batch
from the task dataset.

Each timed update encodes all frames in the window, applies SIGReg to their
encodings, and computes the prediction loss. Timing includes backpropagation,
gradient clipping, and the AdamW step. For the adaptive methods, it also covers
detached discrepancy accumulation, the per-step EMA, and coefficient
updates every ten successful steps. Data loading, transfer, logging, and
epoch-end diagnostics fall outside the timed update.

In each of three repetitions, we run 20 warmup updates and time the next
50. We report the mean and sample standard deviation of the three
repetition means in ms/update. For memory, we report the largest peak
in allocated GPU memory across repetitions, in GiB.

\FloatBarrier
\section{Additional Diagnostics}
\label{app:additional_results}

\subsection{Success--Failure Ranking}
\label{app:success-failure-ranking}

\begin{figure}[tbp]
\centering
\includegraphics[width=\linewidth]{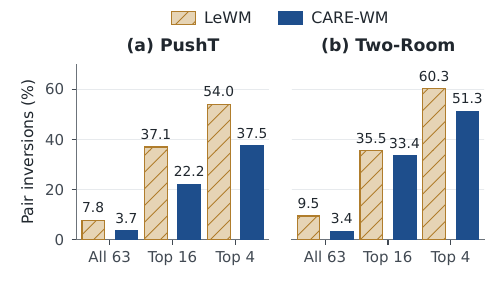}
\caption{
    \textbf{Success--failure ranking within candidate shortlists.}
    An inversion occurs when a failed candidate has strictly lower predicted
    cost than a successful candidate.
    Each model selects its own lowest-cost shortlist; a start is included
    only if both models' shortlists each contain successes and failures.
    Bars show the fraction of inverted success--failure pairs within each
    start ($\%$, $\downarrow$), averaged equally over these common eligible starts.
    For shortlist sizes 63, 16, and 4, respectively, eligible counts are
    2,047, 1,414, and 106 on PushT, and 2,048, 177, and 39 on Two-Room.
}
\label{fig:latent-cost-inversions}
\end{figure}

\Cref{fig:latent-cost-inversions} shows lower mean success--failure
inversion rates for \name{} than for LeWM at every evaluated shortlist
size on both tasks.
Even so, \name{}'s top-four rates remain $37.5\%$ on PushT and $51.3\%$
on Two-Room, based on only 106 and 39 eligible starts, respectively.
The eligible sets differ across shortlist sizes, so these conditional
comparisons do not establish how narrowing a shortlist changes ranking
quality on a fixed population or directly measure closed-loop planning success.
Sampling, execution, and eligibility details are given in
\cref{app:latent-cost-geometry-protocol}.

\subsection{Generated-Context Diagnostics}
\label{app:context_diagnostics}

We compare next-step prediction errors under observed and generated
contexts in recursive rollouts and in SHSC's two-call setting. Following
\cref{app:context_probe_protocol}, we report results for fixed
checkpoints of LeWM, SHSC, and \name{} on Two-Room and PushT. The
three models share 64 windows per task, each from a different episode.

\subsubsection{Aligned-Context Error Decomposition}
\label{app:context_decomposition}

For a fixed model, we pair an observed context $c$ with a generated
context $\hat c$ of the same length and at the same time positions. With
the same actions $\mathbf a$ and target latent $z_+$, the observed-context
residual $r$ and prediction change $d$ are
\begin{equation}
    \begin{aligned}
        r&=\pred_\phi(c,\mathbf a)-z_+,\\
        d&=\pred_\phi(\hat c,\mathbf a)-\pred_\phi(c,\mathbf a).
    \end{aligned}
    \label{eq:context_error_terms}
\end{equation}
The generated-context error is $r+d$, with the exact decomposition
\begin{equation}
    \|r+d\|_2^2=\|r\|_2^2+\|d\|_2^2+2r^\top d.
    \label{eq:context_error_decomposition}
\end{equation}
The sign of the cross term indicates whether the prediction change
reinforces the observed-context residual or partly offsets it. The
magnitude of $d$ depends on both the predictor and the generated inputs
that each model supplies.

\subsubsection{Recursive Generated Contexts}

\Cref{tab:feedback-ablation} reports total MSE for the sixth rollout
prediction after five recursive context updates. On both tasks, SHSC
has lower mean MSE than SHSC (detached), which stops the student
gradient at the teacher-generated inputs during training. BPTT uses a
longer training rollout than SHSC. Its mean MSE is close to SHSC's on
Two-Room and higher on PushT.

The decomposition in \cref{tab:aligned-contexts} shows which terms
account for SHSC's lower mean MSE relative to LeWM at the same horizon.
On Two-Room, SHSC has a larger observed-context error, but its response
term is less than half of LeWM's.
On average, the prediction change partly offsets SHSC's residual
and reinforces LeWM's. On PushT, the two models have similar
observed-context errors. SHSC's smaller response term accounts for its
lower total error despite its larger cross term.
Relative to SHSC, \name{} has smaller residual and response terms on
PushT and slightly higher total MSE on Two-Room.

\begin{table}[tbp]
\centering
\footnotesize
\caption{
    \textbf{Generated-context error decomposition at depth ${h=5}$}
    ($10^{-3}$ MSE units).
    Each term of Eq.~\eqref{eq:context_error_decomposition} is divided by
    $D=192$ and averaged over 64 paired windows per task. The residual,
    response, and signed cross terms sum to the total, up to rounding.
    \textbf{Bold} and \underline{underline} mark the lowest and second-lowest
    residual, response, and total per task; the shaded row is \mbox{\name{}}.
}
\label{tab:aligned-contexts}
\setlength{\tabcolsep}{3pt}
\renewcommand{\arraystretch}{1.1}
\begin{tabularx}{\linewidth}{l*{4}{C}}
\toprule
& \textbf{Residual} & \textbf{Response} & \textbf{Cross} & \textbf{Total} \\
\textbf{Method} & $\|r\|_2^2$ & $\|d\|_2^2$ & $2r^\top d$ & $\|r+d\|_2^2$ \\
\midrule
\multicolumn{5}{l}{\textit{Two-Room}} \\
LeWM & \textbf{1.116} & \phz6.618 & \phm0.872 & \phz8.606 \\
SHSC & 1.737 & \phz\textbf{2.736} & $-$0.141 & \phz\textbf{4.332} \\
\oursrow
\textbf{\name{}} & \underline{1.141} & \phz\underline{2.985} & \phm0.277
    & \phz\underline{4.402} \\
\midrule
\multicolumn{5}{l}{\textit{PushT}} \\
LeWM & 1.351 & 35.973 & $-$0.048 & 37.277 \\
SHSC & \underline{1.322} & \underline{23.542} & \phm0.270 & \underline{25.134} \\
\oursrow
\textbf{\name{}} & \textbf{0.558} & \phz\textbf{6.061} & \phm0.561
    & \phz\textbf{7.180} \\
\bottomrule
\end{tabularx}
\end{table}

\Cref{tab:context-shallow} shows that after one update, SHSC and LeWM
have nearly identical raw MSE on Two-Room, while SHSC has lower raw
MSE on PushT. After two updates, SHSC has lower raw and
variance-normalized mean MSE than LeWM on both tasks.

\begin{table}[tbp]
\centering
\footnotesize
\caption{
    \textbf{Prediction error after $h$ recursive context updates}
    ($10^{-3}$ units, $\downarrow$).
    For each model and task, we encode the six frames following the
    initial context in each of the 64 windows. We compute the population
    variance of each latent coordinate across these 384 encodings and
    average the 192 variances to obtain $v$.
}
\label{tab:context-shallow}
\setlength{\tabcolsep}{3pt}
\renewcommand{\arraystretch}{1.1}
\begin{tabularx}{\linewidth}{l*{4}{C}}
\toprule
& \multicolumn{2}{c}{$h=1$} & \multicolumn{2}{c}{$h=2$} \\
\cmidrule(lr){2-3}\cmidrule(lr){4-5}
\textbf{Method} & MSE & MSE/$v$ & MSE & MSE/$v$ \\
\midrule
\multicolumn{5}{l}{\textit{Two-Room}} \\
LeWM & 2.616 & \underline{2.634} & 4.320 & 4.350 \\
SHSC & \underline{2.611} & 2.636 & \underline{3.606} & \underline{3.640} \\
\oursrow
\textbf{\name{}} & \textbf{1.794} & \textbf{1.854} & \textbf{2.720}
    & \textbf{2.811} \\
\midrule
\multicolumn{5}{l}{\textit{PushT}} \\
LeWM & 2.607 & 2.569 & 4.448 & 4.384 \\
SHSC & \underline{2.219} & \underline{2.269} & \underline{4.180}
    & \underline{4.274} \\
\oursrow
\textbf{\name{}} & \textbf{1.026} & \textbf{1.401} & \textbf{1.754}
    & \textbf{2.395} \\
\bottomrule
\end{tabularx}
\end{table}

\subsubsection{Two-Call Inputs and History Shortening}

\Cref{tab:context-two-call} compares three inputs at the two target
positions of the student call in Eq.~\eqref{eq:student_path}. Full uses
the observed teacher prefix. Short drops the oldest observation.
Generated replaces the remaining observed latents with each model's own
teacher predictions at the same positions, keeping the actions fixed.
Subtracting the Short MSE from the Generated MSE cancels the residual
term in Eq.~\eqref{eq:context_error_decomposition}. The difference is
the sum of the response and signed cross terms.

\begin{table}[tbp]
\centering
\footnotesize
\caption{
    \textbf{Two-call inputs and history shortening}
    ($10^{-3}$ MSE units, $\downarrow$).
    Inputs are full observed (Full), shortened observed (Short), or
    teacher-generated (Gen.). $S_{\mathrm{all}}$ averages the errors at
    $t+2$ and $t+3$, and $S_{\mathrm{last}}$ uses $t+3$ alone.
    Gen./$v$ uses the same variance $v$ as in \cref{tab:context-shallow}.
}
\label{tab:context-two-call}
\label{tab:context-shortening}
\setlength{\tabcolsep}{1.1pt}
\renewcommand{\arraystretch}{1.1}
\begin{tabularx}{\linewidth}{l*{8}{C}}
\toprule
& \multicolumn{4}{c}{$S_{\mathrm{all}}$} & \multicolumn{4}{c}{$S_{\mathrm{last}}$} \\
\cmidrule(lr){2-5}\cmidrule(lr){6-9}
\textbf{Method} & Full & Short & Gen. & Gen./$v$ & Full & Short & Gen. & Gen./$v$ \\
\midrule
\multicolumn{9}{l}{\textit{Two-Room}} \\
LeWM & \underline{1.763} & \underline{1.908} & 3.421 & 3.445
    & \underline{1.994} & \underline{2.131} & 4.075 & 4.103 \\
SHSC & 2.169 & 2.200 & \underline{3.066} & \underline{3.095}
    & 2.486 & 2.477 & \underline{3.729} & \underline{3.764} \\
\oursrow
\textbf{\name{}} & \textbf{1.119} & \textbf{1.057} & \textbf{1.532} & \textbf{1.583}
    & \textbf{1.202} & \textbf{1.140} & \textbf{1.624} & \textbf{1.679} \\
\midrule
\multicolumn{9}{l}{\textit{PushT}} \\
LeWM & \underline{1.317} & \underline{1.409} & 2.615 & 2.577
    & 1.473 & 1.483 & \underline{2.280} & \underline{2.247} \\
SHSC & 1.386 & 1.483 & \underline{2.421} & \underline{2.476}
    & \underline{1.400} & \underline{1.392} & 2.330 & 2.382 \\
\oursrow
\textbf{\name{}} & \textbf{0.602} & \textbf{0.641} & \textbf{0.960} & \textbf{1.311}
    & \textbf{0.615} & \textbf{0.613} & \textbf{0.919} & \textbf{1.256} \\
\bottomrule
\end{tabularx}
\end{table}

For SHSC, dropping the oldest observation slightly lowers final-position
MSE on both tasks. In every row of \cref{tab:context-two-call},
replacing the retained observations with teacher predictions raises
mean MSE by more than the absolute change from shortening the history.

With generated inputs, SHSC has lower mean MSE than LeWM over the two
target positions on both tasks. At the final position on PushT, its
mean MSE is $2.2\%$ higher than LeWM's.

\FloatBarrier
\subsection{Training Dynamics with Detached Feedback}
\label{app:detached-training-dynamics}

Between epochs 10 and 50 of a single SHSC (detached) run on PushT,
the teacher and student prediction losses both decrease on the validation
set. \Cref{tab:detached-training-dynamics} shows that one-step rollout
MSE also decreases. At step five, MSE falls in 21 of the 32 windows,
yet its mean rises from $28.492\times10^{-3}$ to $51.519\times10^{-3}$.
At epoch 50, the three largest terminal errors account for $67.4\%$ of
the sum across all 32 windows.

When each step uses the corresponding observed latents as inputs,
mean fifth-step MSE decreases between checkpoints. At epoch 50 it is
$93.3\%$ lower than when the predictor reuses its own outputs. This
gap is consistent with errors accumulating during recursive prediction.

\begin{table}[htbp]
\centering
\footnotesize
\caption{
    \textbf{Training dynamics of SHSC (detached) on PushT.}
    All values are in $10^{-3}$ MSE units. Both checkpoints use the same
    32 rollout windows from distinct training episodes, with targets
    from their respective encoders. Change is relative to epoch 10.
}
\label{tab:detached-training-dynamics}
\setlength{\tabcolsep}{4pt}
\renewcommand{\arraystretch}{1.1}
\begin{tabularx}{\linewidth}{l*{3}{C}}
\toprule
\textbf{Metric} & \textbf{Epoch 10} & \textbf{Epoch 50} & \textbf{Change} \\
\midrule
\multicolumn{4}{l}{\textit{Validation loss}} \\
Teacher & \phz4.762 & \phz3.117 & $-$34.5\% \\
Student & 10.161 & \phz5.649 & $-$44.4\% \\
\midrule
\multicolumn{4}{l}{\textit{Rollout MSE}} \\
Step 1 & \phz3.319 & \phz1.910 & $-$42.5\% \\
Step 5, generated context & 28.492 & 51.519 & $+$80.8\% \\
Step 5, observed context & \phz4.413 & \phz3.451 & $-$21.8\% \\
\bottomrule
\end{tabularx}
\end{table}

\FloatBarrier
\subsection{SIGReg Gradient Diagnostics}
\label{app:ca-diagnostics}
\label{app:persistent-pressure}

\paragraph{Scale derivative at the Gaussian target.}
With the quadrature used in training, Eq.~\eqref{eq:app_scale_gradient}
predicts a scale derivative of approximately $0.965$ for the expected
raw SIGReg penalty at $\sigma=1$. For Gaussian batches of size 128,
the probe in \cref{app:sigreg_probe_protocol} gives a mean derivative
of $0.952$ with a standard error of $0.126$ over 64 independent repeats.
A small decrease from unit scale therefore lowers the expected penalty.

\paragraph{Encoder gradients with a fixed SIGReg coefficient.}
During SHSC training on Two-Room with $\lambda=0.09$, we measure the
observation-encoder gradients of the normalized teacher--student prediction
loss and the weighted SIGReg penalty before clipping and the AdamW update.
Their median norms are nearly equal in epochs 1--5. By epochs 26--50,
the median prediction norm has fallen to $0.150$ and the median
weighted SIGReg norm to $0.210$.
Across 2,569 late-stage measurements, the ratio of the weighted SIGReg
norm to the prediction norm has a median of $1.398$. The positive
scale derivative and the late-stage norm ratio motivate adapting
$\lambda$ to the calibrated discrepancy in \cref{sec:ca_reg}.

\subsection{Planning Success with Additional Training Seeds}
\label{app:additional-training-seeds}

\Cref{tab:additional-training-seeds} reports planning success after
retraining \name{} on PushT and Two-Room with seeds 3073 and 3074.
On each task, both training runs yield mean success rates within
three percentage points of the corresponding \name{} result in
\cref{tab:planning}.

\begingroup
\setlength{\intextsep}{6pt}
\begin{table}[H]
\centering
\footnotesize
\captionsetup{skip=3pt}
\caption{
    \textbf{\name{} planning success (\%) by training seed.}
    Each entry reports the mean and sample standard deviation across
    six evaluation seeds for one trained model, with 50 episodes per
    evaluation seed.
}
\label{tab:additional-training-seeds}
\setlength{\tabcolsep}{4pt}
\renewcommand{\arraystretch}{1.1}
\begin{tabularx}{\linewidth}{l*{2}{C}}
\toprule
\textbf{Task} & \textbf{Training seed 3073} & \textbf{Training seed 3074} \\
\midrule
PushT & \res{96.67}{2.07} & \res{95.33}{2.73} \\
Two-Room & \res{93.33}{3.01} & \res{94.33}{3.20} \\
\bottomrule
\end{tabularx}
\end{table}
\endgroup

\end{document}